\documentclass[conference]{IEEEtran}
\IEEEoverridecommandlockouts
\usepackage{cite}
\usepackage{amsmath,amssymb,amsfonts}
\usepackage{algorithmic}
\usepackage{graphicx}
\usepackage{textcomp}
\usepackage{xcolor}
\usepackage{glossaries}
\usepackage{textcomp}
\usepackage{svg}
\usepackage[binary-units=true]{siunitx}
\usepackage{booktabs}
\newcommand{\ra}[1]{\renewcommand{\arraystretch}{#1}}

\def\BibTeX{{\rm B\kern-.05em{\sc i\kern-.025em b}\kern-.08em
    T\kern-.1667em\lower.7ex\hbox{E}\kern-.125emX}}

\newacronym{cnn}{CNN}{convolutional neural network}
\newacronym{map}{mAP}{mean average precision}
\newacronym{qat}{QAT}{quantization aware training}
\newacronym{gpu}{GPU}{graphics processing unit}
\newacronym{cpu}{CPU}{central processing unit}
\newacronym{fpga}{FPGA}{field programmable gate array}
\newacronym{asic}{ASIC}{application-specific integrated circuit}
\newacronym{soc}{SoC}{system on chip}
\newacronym{yolo}{YOLO}{you only look once}
\newacronym{rcnn}{R-CNN}{region-based convolutional neural network}
\newacronym{ssd}{SSD}{single shot detector}
\newacronym{mcu}{$\mu$C}{microcontroller unit}
\newacronym{ml}{ML}{machine learning}
\newacronym{ai}{AI}{artificial intelligence}
\newacronym{dnn}{DNN}{deep neural network}
\newacronym{bbox}{Bbox}{bounding box}

\usepackage{eso-pic}
\usepackage{url}
\AddToShipoutPictureBG*{
  \AtPageUpperLeft{%
    \put(0,-40){\raisebox{15pt}{\makebox[\paperwidth]{\begin{minipage}{21cm}\centering
      \textcolor{gray}{This paper has been accepted for publication at the \\ IEEE Conference on Artificial Intelligence Circuits and Systems (AICAS), Hangzhou, 2023.}
    \end{minipage}}}}%
  }
  \AtPageLowerLeft{%
    \raisebox{25pt}{\makebox[\paperwidth]{\begin{minipage}{21cm}\centering
      \textcolor{gray}{ \copyright 2023 IEEE.  Personal use of this material is permitted.  Permission from IEEE must be obtained for all other uses, in any current or future media, including reprinting/republishing this material for advertising or promotional purposes, creating new collective works, for resale or redistribution to servers or lists, or reuse of any copyrighted component of this work in other works.
      }
    \end{minipage}}}%
  }
}

\begin{document}

\title{TinyissimoYOLO: A Quantized, Low-Memory Footprint, TinyML Object Detection Network for Low Power Microcontrollers  \\
\thanks{The authors would like to thank \textit{armasuisse Science \& Technology} for funding this research.}
}

\author{\IEEEauthorblockN{Julian Moosmann, Marco Giordano, Christian Vogt, Michele Magno}
\IEEEauthorblockA{\textit{Center for Project Based Learning - ETH Zürich}\\
julian.moosmann, marco.giordano, christian.vogt, michele.magno@pbl.ee.ethz.ch}
}


\maketitle

\begin{abstract}
This paper introduces a highly flexible, quantized,  memory-efficient, and ultra-lightweight object detection network, called TinyissimoYOLO. It aims to enable object detection on microcontrollers in the power domain of milliwatts, with less than $\SI{0.5}{\text{MB}}$ memory available for storing \gls{cnn} weights. The proposed quantized network architecture with $\SI{422}{\text{k}}$ parameters, enables real-time object detection on embedded microcontrollers, and it has been evaluated to exploit \gls{cnn} accelerators. In particular, the proposed network has been deployed on the MAX78000 microcontroller achieving high frame-rate of up to $\SI{180}{\textbf{fps}}$ and an ultra-low energy consumption of only $\SI{196}{\mu\joule}$ per inference with an inference efficiency of more than $\SI{106}{\text{MAC/Cycle}}$. TinyissimoYOLO can be trained for any multi-object detection. However, considering the small network size, adding object detection classes will increase the size and memory consumption of the network, thus object detection with up to 3 classes is demonstrated. Furthermore, the network is trained using quantization-aware training and deployed with 8-bit quantization on different microcontrollers, such as STM32H7A3, STM32L4R9, Apollo4b and on the MAX78000's \gls{cnn} accelerator. Performance evaluations are presented in this paper. 
\end{abstract}
\glsreset{cnn}
\glsreset{map}


\vspace{-0.35cm}
\section{Introduction}
Object detection on edge devices such as \glspl{mcu} have the capability of reducing detection latency and increasing the overall energy efficiency by running on-device network inferences \cite{dutta_tinyml_2021, boner2022tiny}. In addition, the sensitive transmission of sensor data is reduced or substituted, reducing privacy issues to a minimum. Furthermore, emerging dedicated hardware accelerators for \gls{ml} models are enabling edge processing and edge \gls{ai} reducing the energy consumption for inference and enabling real-time on-board processing on resource-constrained microcontrollers\cite{merenda_edge_2020}. On such constrained devices, computational power is significantly reduced and does not allow for the deployment of classical object detection \glspl{dnn} such as \gls{yolo} \cite{redmon_you_2016}, \glspl{rcnn} \cite{ren_faster_2015} or \glspl{ssd} \cite{liu_ssd_2016} because their memory requirements are significantly exceeding the available memory of few kilobytes typically available in such devices. However, their fundamental ideas are of importance for designing new \glspl{dnn}, especially for the emerging field of edge \gls{ai}.

In fact, to keep the power consumption in the order of a few milliwatts, the computational power on \glspl{mcu} is significantly lower compared to CPUs and GPUs. Thus, object detection networks need to be carefully designed and optimized for such tiny devices in particular to achieve a small memory footprint and a high number of operations processed per cycle \cite{wang2020fann}. To overcome the memory constraints, several different methods have been recently reported in literature, such as pruning \cite{zhu_prune_2017}, \cite{molchanov_pruning_2017}, quantization \cite{gholami_survey_2021}, \cite{han_deep_2016}, new frameworks developed for memory efficient inference \cite{lin_mcunet_2020}, patch-based inference scheduling \cite{lin_mcunetv2_2021} or neural architecture searches including search spaces specialized for memory-constrained devices \cite{cai_once-for-all_2020}, \cite{cai_-device_2019}. These techniques are used to reduce the memory and complexity of existing state-of-the-art networks to deploy on a tiny device while keeping the original network's structure and still achieving similar inference accuracy while performance (especially inference speed) is often neglected \cite{scherer2021tinyradarnn}. Among other techniques, quantization is one of the most promising and popular, as it reduces both the memory requirements and increases the number of operations per second that \glspl{mcu} can perform. Li et al. (2019) \cite{li_fully_2019} have shown that quantizing an object detection network to 4-bit, they achieve state-of-the-art prediction accuracy with a \gls{map} loss of only $\SI{2}{\percent}$ to $\SI{5}{\percent}$ compared to its full precision counterpart while having 8x less memory occupation. \\ To the best of our knowledge, there is still no previous work that adopts those techniques to achieve a generalized object detection network, ready for deployment on edge devices and \glspl{mcu} with less than $\SI{0.5}{\text{MB}}$ of weight memory, that are able to achieve similar accuracy performance of larger networks.

This paper proposes a quantized and highly accurate object detection \gls{cnn} based on the architecture of \gls{yolo} \cite{redmon_you_2016} suitable for edge processors with limited memory and computational resources. The proposed network is composed of quantized convolutional layers with 3x3 kernels and a fully connected layer at the output. It is designed for having a low memory footprint of less than $\SI{0.5}{\text{MB}}$. The proposed network is trained and evaluated on the WiderFace dataset \cite{yang_wider_2016}. Furthermore, to showcase multi-object detection capability, while keeping the network small, it has been trained and evaluated on a sub-set of the PascalVOC \cite{everingham_pascal_2015} dataset (3 out of the 20 classes, namely: person, chair and car). Finally, the network is deployed quantized and memory-efficient on different \gls{mcu} architectures, such as the STM32H7A3 and STM32L4R9 from STMicroelectronics, Ambiq's Apollo4b and on a novel microcontroller, MAX78000 from Analog Devices, which has a built-in \gls{cnn} accelerator. The performance of the different architectures is compared and it will be shown, how the MAX78000 outperforms the other \glspl{mcu}. Furthermore, this paper investigates the effect of \gls{map} against relative object size dimensions within images. This evaluation helps to understand which object size should be chosen when training the generalized object detection network.

\section{TinyissimoYOLO}
\begin{figure*}[htbp]
\centerline{\includegraphics[width=0.9\textwidth]{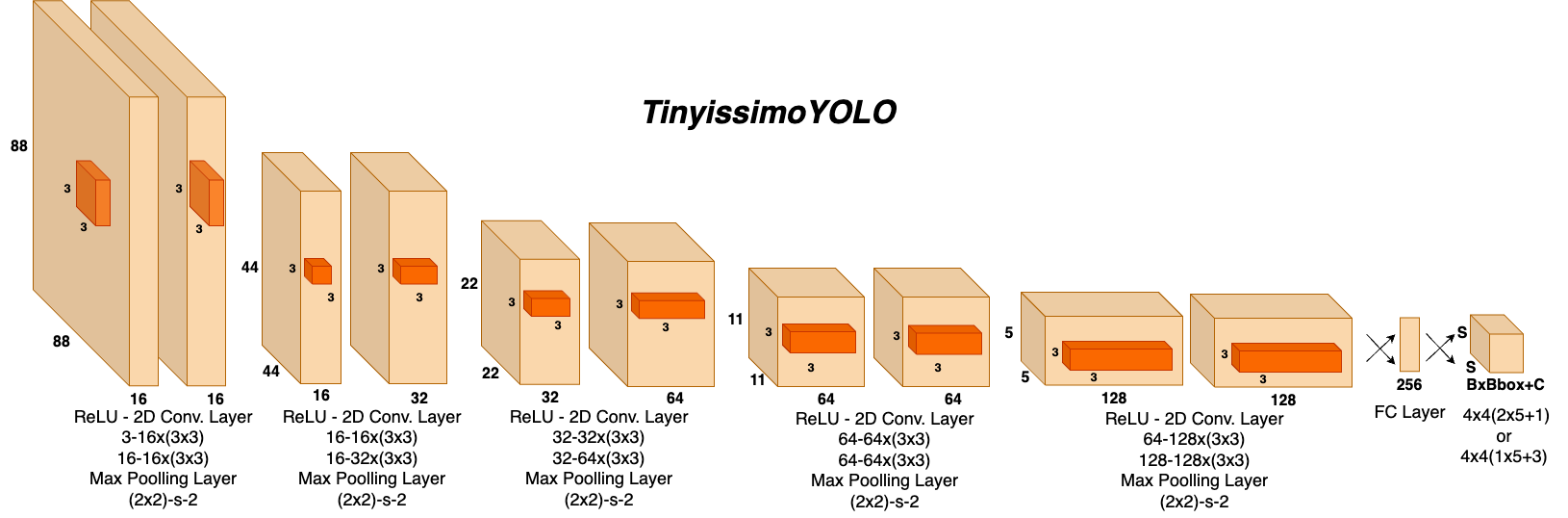}}
\caption{TinyissimoYOLO proposed by this paper.}
\label{fig:tinyissimoYOLO}
\end{figure*}

The ultra-lightweight object detection network designed in this paper for operation on \gls{mcu}, named TinyissimoYOLO, is a general object detection network and can be seen in Figure \ref{fig:tinyissimoYOLO}. It can be deployed on any hardware with at least $\SI{0.5}{\text{MB}}$ of network parameter memory, i.e. flash, available. Not only is it capable of being deployed on any ARM Cortex-M microcontrollers, but it can also be deployed on \glspl{mcu} with built-in hardware accelerators, such as the MAX78000 from Analog Devices, GAP9 \cite{reuther_survey_2020} from Greenwaves, Sony' Spresense \cite{noauthor_overview_nodate-1} or Syntiant TinyML \cite{yousefi_intelligence_2019}, among others. The network can be trained on any object detection class and supports multi-class object detection with the cost of minimal increasing memory.

To facilitate small size and to be deployable in an efficient way on all existing microcontrollers, as well as exploiting  \gls{cnn} emerging hardware accelerators that are starting to be embedded in recent microcontrollers, TinyissimoYOLO has been designed to consist of only convolutional layer-operations and fully connected linear layers, which are largely optimized in hardware and software toolchains. As we mentioned, the goal of the design is to reach outstanding accuracy performance with a memory envelope of $\SI{0.5}{\text{MB}}$ and with a minimal number of operations and yet a high number of operations per cycle.

\subsection{Network design}
TinyissimoYOLO is inspired by the \gls{yolo}v1 \cite{redmon_you_2016} architecture. Major design decisions of the TinyissimoYOLO network encompass the following 4 points:
\begin{itemize}
    \item input image size,
    \item hidden convolutional layers,
    \item fully connected linear layer before the output layer,
    \item output layer size.
\end{itemize}
\textbf{Input image size} has been chosen such that the network runs on all the previously mentioned \glspl{mcu}. Limiting factor is the \gls{cnn} accelerator of the MAX78000, which does not support \gls{cnn} inputs greater than 90x91 without using a specialized mode. Thus the input of 88x88 is chosen because it is a tread-off between maximizing the image size and being able to do pooling on the input dimensions without rounding the dimensions down to the 4th pooling layer. The network's \textbf{hidden convolutional layer} parameter memory scales by the number of input channels times the number of output channels times the filter size. Even though in the original YOLOv1 network high channel count layers are preferred, 
 the small memory size on \glspl{mcu} and accelerators will not allow such layers. For example, one convolutional layer with 256 input and output channels and a kernel of 3x3, consumes already more than $\SI{0.5}{\text{MB}}$ of quantized 8-bit weights. To ensure staying below $\SI{0.5}{\text{MB}}$ of network memory, only convolutional layers with 3x3 filters are used, since it scales the weight memory only by a factor of 9x, instead of 25x (for 5x5 kernel) or 49x (for a 7x7 kernel). Even though it would be possible to use bigger filter sizes (e.g. \gls{yolo}v1 used a filter of 7x7 in the input) we focused on using convolutional layers with 3x3 filters and keeping the network as simple as possible. Furthermore, we increase the number of channels with increasing network depth, starting with 16 output channels in the \gls{cnn}'s input layer and ending with 128 output channels at the \gls{cnn}'s output. Max pooling with a stride of two decreases the channel resolution every second layer. 
 Thereafter, the fully connected layer predicts the output probabilities and coordinates out of the features extracted by the former convolutional layers. As such, the \textbf{fully connected linear layer which connects the \gls{cnn} to the output layer}, is chosen to be 256 and therefore is larger than the output layer size of 4x4(2xBbox+C) for mapping the \gls{cnn}'s features onto the regression, classification output.
The network's \textbf{output layer} uses the \gls{yolo}v1 output of the size SxS(BxBbox+C). Where SxS is the grid, within each cell B \gls{bbox} (with the objects width, height, centre in x- and y-axis, relative to the grid cell size and prediction certainty) of possible detections are calculated by the network as well as the C possibly detected classes. As the network's input image size is set to 88x88 pixels a smaller output grid size of 4x4 and $B=2$ is used. The value of $B$ was found as a trade-off between significantly more memory consumption and having less prediction accuracy. $B=3$, increased the output layer memory consumption by a factor of 1.5x but did not increased the \gls{map} significantly. Therefore, $B=2$ has been used for training the network on WiderFace. To ensure deployability on the \gls{cnn} accelerator of MAX78000 and training a multi-object detection network for 3 classes of PascalVOC, $B=1$ has been chosen. The single-class TinyissimoYOLO with $B=2$ is $\SI{106}{\percent}$ bigger than the multi-class TinyissimoYOLO with $B=1$. 
For training the TinyissimoYOLO network, we use the YOLO loss function introduced in the YOLOv1 paper \cite{redmon_you_2016} by Redmon et al. (2016) to train the network. Evaluation is done by using the \gls{map} as it has been used by Everingham et al. (2015) \cite{everingham_pascal_2015}.

\subsection{Training and Dataset}
For training, testing and validation of the TinyissimoYOLO network, the WiderFace dataset  \cite{yang_wider_2016} was used. $\SI{90}{\percent}$ of WiderFace training dataset was used to train the network, while $\SI{10}{\percent}$ was used for validation. Testing of the network was done with the corresponding test dataset of WiderFace. Due to TinyissimoYOLO's small input size of 88x88x3, and the 4x4 network output grid predictions, we evaluated the network's accuracy by restricting the training data to images containing less or equal to 10 and 5 objects. As such, the objects are ensured to be visible in the downscaled 88x88 pixel images. With the restriction made on the training dataset of WiderFace, we evaluate the influence of the number of objects within an image on \gls{map}. Additionally, for showing the network's ability for general multi-object detection, the network is trained and evaluated on the PascalVOC dataset \cite{everingham_pascal_2015} with a restriction of maximal three objects per image, due to the increased difficulty of the multi-class object detection problem. To not only deploy the network on the STM32s and Apollo4b but also to fit in the MAX78000 \gls{cnn} accelerator's memory, the multi-object TinyissimoYOLO is trained and evaluated using only 3 of the 20 classes, namely person, chair and car and thus stays bellow the sub-$\SI{0.5}{\text{MB}}$ memory size restriction set by us.
The 3 classes were chosen because of their object occurrences within the dataset and thus to train the network with a balanced dataset. Deploying the network on real hardware, experimental tests can be conducted with many different sceneries inside and outside of buildings. \\
Because the network is designed to be deployable on \glspl{mcu} and on \gls{cnn} accelerators in the most efficient way, the network is quantized from 32-floating point to 8-bit integers. This further reduces computation cycles and memory requirements. In order to optimize the network's fine-tuning, the training is already conducted aware of this quantization. 
The network is trained 350 epochs with floating-point precision and thereafter switches to another 300 epochs trained quantization-aware. The starting learning rate as well as the weight decay is set to be $5*10^{-4}$. The used optimizer is SGD \cite{saad_-line_1999}. The learning rate is scheduled with a multi-step learning rate for improved learning speed \cite{goyal_accurate_2018}.

\subsection{$\mu$C implementation}
By experimental evaluation, we demonstrate that TinyissimoYOLO is efficiently deployed on several low-power devices, such as STM32L4R9 with an ARM Cortex-M4F core, STM32H7 with an ARM Cortex M7 core, Apollo4b the world's more energy-efficient ARM Cortex-M4F core, and MAX78000 (using its built-in \gls{cnn} accelerator). Deploying the designed and quantized network on the \glspl{mcu}'s ARM Cortex M cores, is straightforward, as we designed the network to be compatible with TensorFlow-lite micro \cite{david_tensorflow_2021}. Next to the \gls{mcu} only deployment, it is shown that the network can also operate well on a \gls{mcu} with a built-in \gls{cnn} accelerator (MAX78000) by deploying it with the provided deployment tools from Analog Devices \cite{noauthor_adi_2022}.
TinyissimoYOLO fits the MAX78000's \gls{cnn} accelerator's limitations. The most important considerations for the MAX78000 are that the allowed network memory footprint of its weights can only be 442kB. Considering the data passed through the architecture, the input size is constrained to $<$90x91 input width and height, when using the non-streaming mode of the \gls{cnn} accelerator. All network operations needed by TinyissimoYOLO are covered by the design of the hardware accelerator and its software deployment tools.
A full pipeline demonstrator with MAX78000FTHR \gls{mcu} and on-device image sensor CameraCubeChip\textregistered is built and used for validating the system with real-life data.

\section{Experimental Results}
The networks have been trained on a subset of the WiderFace \cite{yang_wider_2016} and the PascalVOC datasets as explained in section \textit{B. Training and Dataset}. They have been evaluated on the whole test dataset of WiderFace as well as on the restricted datasets. Table \ref{tab:map_results_wf} 
lists the evaluation of the trained TinyissimoYOLO networks and compares the \gls{map} \cite{everingham_pascal_2015} when trained with a different number of max. objects allowed per image. On WiderFace, TinyissimoYOLO achieves a \gls{map} \cite{everingham_pascal_2015} of $\SI{45}{\percent}$ when evaluated on the whole dataset and up to $\SI{73}{\percent}$ \gls{map} when restricting the evaluation dataset of the network to max. 5 objects per image. Restricting the dataset during training to max. 10 or 5 objects per image, a \gls{map} of $\SI{46}{\percent}$ and $\SI{44}{\percent}$ \gls{map} is achieved when evaluating on the whole dataset. Evaluating the restricted trained networks on the restricted datasets up to $\SI{75}{\percent}$ \gls{map} is achieved with an increase of $\SI{2}{\percent}$ compared to training the network on the whole dataset. The TinyissimoYOLO network trained and evaluated on the PascalVOC dataset for the object detection classes person, chair and car achieves \gls{map} of up to $\SI{57}{\percent}$, see Table \ref{tab:map_results_voc}. Here a global restriction to maximal three objects per image is made because of the increased difficulty of the multi-class object detection problem. 
\vspace{-0.55cm}
\begin{table}[htbp]
    \ra{1.1}
    \caption{TinyissimoYOLO network trained and evaluated on WiderFace (face detection). Maximal allowed objects per image have been restricted for training and evaluation.}
    \begin{center}
    \begin{tabular}{@{}llll@{}}\toprule
         & \multicolumn{3}{c}{\textbf{\gls{map}}} \\
        \cmidrule{2-4}
        \multicolumn{1}{l}{\textbf{WiderFace}} & \textbf{no restriction} &  \textbf{max 10 obj.} & \textbf{max 5 obj.} \\
        \midrule
        \textbf{no restriction} & $\SI{45.3}{\percent}$ & $\SI{64.1}{\percent}$ & $\SI{73.5}{\percent}$ \\
        \textbf{max 10 obj.}  & $\SI{46.2}{\percent}$ & $\SI{65.7}{\percent}$ & $\SI{75.4}{\percent}$ \\
        \textbf{max 5 obj.}   & $\SI{43.5}{\percent}$ & $\SI{62.4}{\percent}$ & $\SI{72.7}{\percent}$ \\
        \bottomrule
        \end{tabular}
    \label{tab:map_results_wf}
    \end{center}
\end{table}
\vspace{-0.8cm}
\begin{table}[htbp]
    \ra{1.1}
    \caption{TinyissimoYOLO network trained and evaluated on PasccalVOC with classes: person, chair and car. Maximal allowed objects per image have been restricted to 3 objects for training and evaluation.}
    \begin{center}
    \begin{tabular}{@{}lllll@{}}
    \toprule
         & \multicolumn{4}{c}{\textbf{\gls{map}}} \\
        \cmidrule{2-5}
        \textbf{Dataset} & \textbf{person} &  \textbf{chair} & \textbf{car} & \textbf{overall} \\
        \midrule
        \textbf{PascalVOC$^{\mathrm{a}}$} & $\SI{57.4}{\percent}$ & $\SI{30.2}{\percent}$  & $\SI{65.1}{\percent}$ & $\SI{58.5}{\percent}$ \\
        \bottomrule
        \multicolumn{5}{c}{$^{\mathrm{a}}$ training dataset with person / chair / car classes} \\
    \end{tabular}
    \label{tab:map_results_voc}
    \end{center}
\end{table}

\vspace{-0.35cm}
Given the TiniyssimoYOLO's performance on the datasets, the TinyissimoYOLO networks (trained on WiderFace and PascalVOC) are now compared when deployed quantized to 8-bits on the different \glspl{mcu}. Since both networks yield the exact same performance results, no network distinction is made for the results on the following metrics: 
\begin{itemize}
\item power efficiency [$\mu$W/MHz],
\item inference efficiency [MAC/Cycles] (which tells how well the device can parallelise the network execution),
\item inference latency [ms],  
\item energy per inference [mJ/Inf.].
\end{itemize}

\begin{figure}[htbp]
\centerline{\includegraphics[width=\columnwidth]{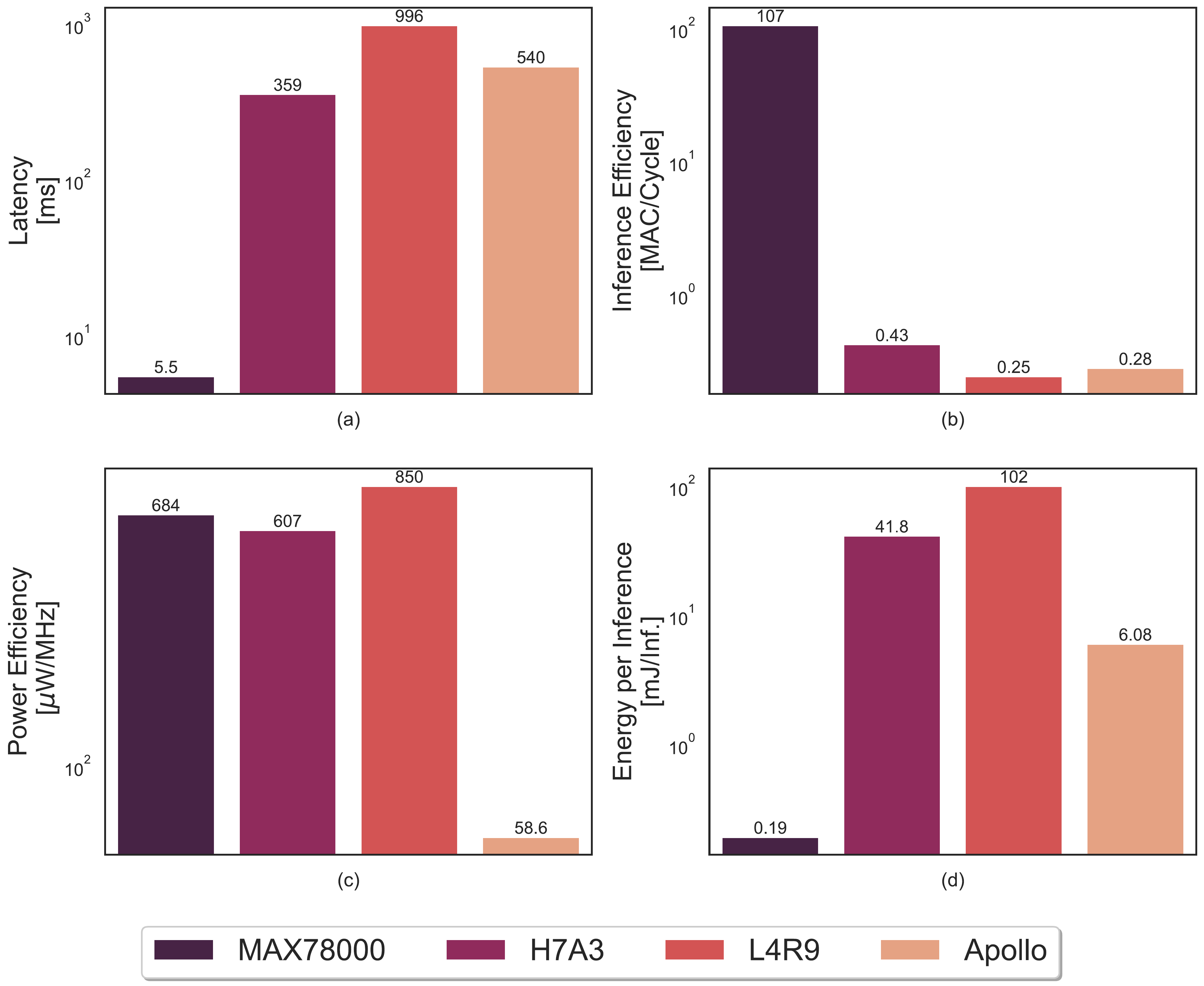}}
\caption{TinyissimoYOLO performance comparison when deployed quantized to 8-bit on the different architectures. The \gls{cnn} accelerated MAX78000 \gls{mcu} outperforms the other architectures in terms of latency, inference efficiency and energy per inference.}
\label{fig:dev_comparison}
\end{figure}

The metrics are measured by deploying TinyissimoYOLO on the different target devices and measuring the power consumption of the \glspl{mcu} only with the following \gls{mcu} configurations: MAX78000 $@$ (\SI{1.2}{\volt}; \SI{50}{\text{MHz}}), STM32H7A3 $@$ (\SI{3.3}{\volt}; \SI{192}{\text{MHz}}), STM32L4R9 $@$ (\SI{3.3}{\volt}; \SI{120}{\text{MHz}}) and Apollo4b $@$ (\SI{1.8}{\volt}; \SI{192}{\text{MHz}}). The metrics are chosen as proposed by Giordano et al. (2022) \cite{giordano_survey_2022} to evaluate the architecture's latency, power and energy efficiency as well as the hardware's capability of parallelising the network's computational workload to its processor(s). Figure \ref{fig:dev_comparison} a), compares the latency of the network being executed on the different \glspl{mcu}. Due to the usage of the custom \gls{cnn} hardware accelerator, MAX78000 runs one inference within \SI{5.5}{\milli\second} while the second-fastest is STM32H7A3 with \SI{359}{\milli\second}. Therefore, MAX78000 outperforms the others by a factor of $>$65x. Figure \ref{fig:dev_comparison} b) shows the inference efficiency of the different architectures. The accelerated \gls{cnn} hardware of MAX78000 showcases its ability to parallelise the \gls{cnn} workload with \SI{107}{\text{MAC/Cycle}} while all the others need at least $2$-\SI{4}{\text{cycles}} to execute one MAC. In \mbox{Figure \ref{fig:dev_comparison} c)}  Apollo4b outperforms all the others by only consuming \SI{59}{\text{$\mu$\watt/MHz}} and thus being the most power efficient \gls{mcu} within this comparison. Figure \ref{fig:dev_comparison} d) shows the energy efficiency of the devices. Despite the fact that the Apollo4b has best-in-class power efficiency, the MAX78000 hardware accelerator's fast and inference-efficient execution manages to be the overall, most energy efficient with only $\SI{196}{\mu\joule}$ per Inference and outperforming by a factor of 32x compared to the second best, being the Apollo4b with \SI{6.1}{\milli\joule} per Inference.
Last but not least the TinyissimoYOLO network consumes \SI{422}{\kilo\byte} of memory when trained on 1 object class, e.g. WiderFace and \SI{398}{\kilo\byte} when trained on 3 object classes, e.g. PascalVOC (person, chair and car). The input image of 88x88x3 pixels consumes an additional \SI{25}{\kilo\byte} of memory space, while \SI{350}{\kilo\byte} of RAM (when deployed on the ARM Cortex-M cores) is sufficient for executing an inference.

\vspace{-0.35cm}
\section{Conclusion}
This work presented TinyissomoYOLO, a multi-object detection network showcased to be used for edge applications with small amount of objects to be detected simultaneously. The network can be deployed on any \gls{mcu} with a minimal required flash of less than $\SI{0.5}{\text{MB}}$ for its network parameters.\\
Evaluations showed, training the network on input images with objects adequate for its input size (restricting the dataset to maximal 10 or 5 objects per image) increases not only network performance on a restricted evaluation but also when evaluating the network on the original validation dataset, achieving up to $\SI{45}{\percent}$ \gls{map} for WiderFace. Evaluating the network on a restricted PascalVOC dataset, the network achieves $\SI{59}{\percent}$ \gls{map} with $\SI{57}{\percent}$ / $\SI{30}{\percent}$ / $\SI{65}{\percent}$ on the classes \mbox{person / chair / car} respectively. 
Comparing low power \gls{mcu} devices when running the TinyissimoYOLO network on those, reveals, the MAX78000's \gls{cnn} accelerated hardware is the most energy efficient among the evaluated devices and is able to achieve fast inference times running in real-time on \glspl{mcu} with speeds up to $\SI{180}{\text{fps}}$.

\bibliographystyle{IEEEtran}
\bibliography{IEEEabrv,AICAS_TinyissimoYOLO}

\vspace{12pt}
\end{document}